\documentclass[letterpaper, 10 pt, conference]{ieeeconf}  

\makeatletter
\let\NAT@parse\undefined
\makeatother

\IEEEoverridecommandlockouts                              

\usepackage{times}  
\usepackage{helvet}  
\usepackage{courier}  
\usepackage{url}  
\usepackage{graphicx}  
\usepackage{balance}
\usepackage{color}
\usepackage{amsmath}
\usepackage{amssymb}
\usepackage{multirow}
\usepackage[font={small}]{caption}
\usepackage[utf8]{inputenc}
\usepackage{booktabs}
\usepackage{dblfloatfix}
\usepackage{tikz}
\usepackage{cite}
\usepackage{algorithm}
\usepackage{algpseudocode}
\usepackage{cancel}
\usepackage{mathtools}
\usepackage{cuted}
\usepackage{siunitx}
\usetikzlibrary{shapes, arrows}
\usetikzlibrary{backgrounds, fit, calc, 3d}
\usetikzlibrary{positioning}
\definecolor{dred}{rgb}{1,0,0}
\definecolor{dblue}{rgb}{0,0,1}

\renewcommand{\baselinestretch}{0.99}

\usepackage{diagbox}

\newcommand{\etal}{\textit{et al. }}

\newcommand\figref[1]{Fig.~{\ref{#1}}}
\newcommand\tabref[1]{Table~{\ref{#1}}}

\newcommand{\ourmodel}{LUMOS}

\usepackage[hidelinks]{hyperref}
\usepackage{xcolor}
\hypersetup{
	colorlinks,
	linkcolor={red!50!black},
	citecolor={blue!50!black},
	urlcolor={blue!80!black}
}

\DeclarePairedDelimiterX{\infdivx}[2]{(}{)}{%
  #1\;\delimsize\|\;#2%
}
\newcommand{\kl}{D_{\text{KL}}\infdivx}

\title{\LARGE \bf LUMOS: Language-Conditioned Imitation Learning with World Models}

\author{
    Iman Nematollahi$^{1}$,
    Branton DeMoss$^{2}$,
    Akshay L Chandra$^{1}$,
    Nick Hawes$^{2}$,
    Wolfram Burgard$^{3}$,
    Ingmar Posner$^{2}$\\
    \thanks{$^{1}$University of Freiburg. $^{2}$University of Oxford. $^{3}$University of Technology Nuremberg.}
}

\begin{document}

\maketitle
\begin{abstract}
    We introduce LUMOS, a language-conditioned multi-task imitation learning framework for robotics. LUMOS learns skills by practicing them over many long-horizon rollouts in the latent space of a learned world model and transfers these skills zero-shot to a real robot. By learning on-policy in the latent space of the learned world model, our algorithm mitigates policy-induced distribution shift which most offline imitation learning methods suffer from. LUMOS learns from unstructured play data with fewer than 1\% hindsight language annotations but is steerable with language commands at test time. We achieve this coherent long-horizon performance by combining latent planning with both image- and language-based hindsight goal relabeling during training, and by optimizing an intrinsic reward defined in the latent space of the world model over multiple time steps, effectively reducing covariate shift. In experiments on the difficult long-horizon CALVIN benchmark, LUMOS outperforms prior learning-based methods with comparable approaches on chained multi-task evaluations. To the best of our knowledge, we are the first to learn a language-conditioned continuous visuomotor control for a real-world robot within an offline world model. Videos, dataset and code are available at \url{http://lumos.cs.uni-freiburg.de}.
\end{abstract}

\section{Introduction}
\label{sec:introduction}
Building systems that can perform long-horizon tasks specified by natural language is a long-standing goal in robotics~\cite{mees2022calvin}. When complete task specifications are available, reinforcement learning (RL) methods can be used to produce policies by trial and error. However, typical RL algorithms can require millions of episodes to learn good policies, which is infeasible for many robotics tasks. Furthermore, many tasks are difficult to specify as reward functions, requiring substantial engineering effort to elicit reasonable learned skills from RL alone.\looseness=-1

Imitation learning offers an alternative by learning directly from task demonstrations. Naïve approaches like behavior cloning~\cite{BC} predict actions from states based on demonstration data and apply them during deployment. However, these methods overlook the sequential nature of decision-making. Since each observation depends on the previous action, the independence assumption in standard statistical machine learning is broken. Ross and Bagnell~\cite{Bagnell1} showed that a behavior-cloned policy with error $\epsilon$ has expected regret of $\mathcal{O}(T^2\epsilon)$, growing quadratically with decision horizon $T$. Intuitively, because the behavior-cloned policy is not trained under its own distribution, small prediction errors compound, pulling the policy out of distribution to unseen states, which is especially problematic for long-horizon tasks due to the quadratic regret growth.\looseness=-1

\begin{figure}[t]
	\centering
	\includegraphics[width=1\columnwidth]{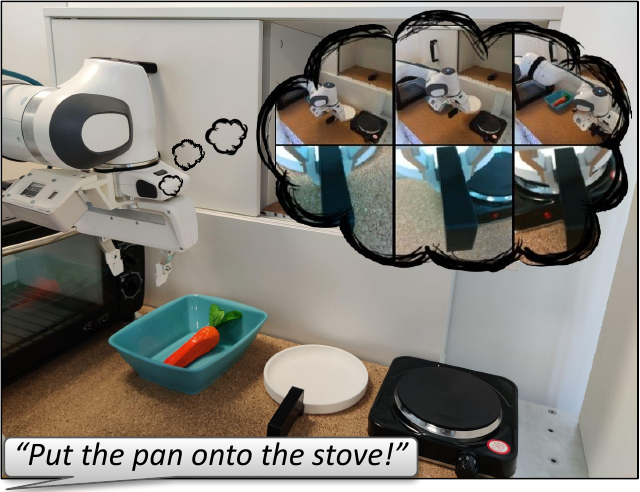}
        \caption{\textbf{\ourmodel}~learns a general-purpose language-conditioned visuomotor policy within the latent space of a learned world model. By optimizing an intrinsic reward to match expert performance, it recovers from its own mistakes across multiple time steps and reduces covariate shift. As a result, LUMOS can handle complex, long-horizon robot tasks from abstract language instructions in real-world scenarios, without online learning or fine-tuning.}
	\label{fig:introduction}
        \vspace{-0.8cm}
\end{figure}

An approach to alleviating distribution shift in off-policy learning is to train on-policy in a simulation of the environment. However, simulators may be unavailable or suffer from the sim2real problem, where policies that perform well in simulation fail to generalize to the real world due to the ``reality gap''—a mismatch between real and simulated dynamics~\cite{Hwangbo_2019}.\looseness=-1

World models offer a promising alternative to hand-crafted simulators~\cite{ha2018world}, approximating dynamics by predicting states or observations conditioned on actions from real trajectories over multiple steps. By compressing observations into latent representations that capture temporal context, contemporary world models~\cite{muzero, hafner2023mastering} achieve high-fidelity predictions over thousands of time steps. Learning within the world model's latent space enables both on-policy and long-horizon learning, mitigating distribution shift and avoiding the reality gap from mis-specified simulators.\looseness=-1

Taking steps towards achieving language-conditioned, long-horizon, multi-task policies, we introduce \textbf{LUMOS}, a language-conditioned imitation learning framework. LUMOS employs world models to learn a multi-task policy that can be guided by free-form natural language instructions entirely offline and transferred to the real world in a zero-shot fashion. It leverages the latent-matching intrinsic reward proposed by DITTO~\cite{demoss2023ditto} within the world model's latent space, enabling its actor-critic agent to recover expert performance without any online interaction. To further enhance the long-horizon performance of the actor-critic agent, we build upon recent advancements in imitation learning~\cite{lynch2020learning,mees2022matters}, and augment LUMOS with a latent planning network. This network uses the benefit of hindsight during training to distill predicted information about the future trajectory into a ``latent plan''. We demonstrate that LUMOS can learn imitation policies solely within the world model's latent space, achieving zero-shot transfer to the real world. Extensive tests on the CALVIN benchmark and in real-world scenarios show that our algorithm excels in handling multi-objective, long-horizon tasks specified in natural language, outperforming previous learning-based methods.\looseness=-1

Our contributions are as follows: 1) \textbf{Improved Long-Horizon Imitation Learning:} \ourmodel~reduces covariate shift and enhances long-horizon performance by combining world-model-based exploration with latent planning, enabling agents to practice skills over simulated long horizons using full demonstration trajectories. Ablation studies show that multi-step practice in latent space significantly outperforms single-step behavioral cloning.
2) \textbf{Language Goal-Conditioned Learning:} Our language-conditioned imitation learning framework outperforms prior methods with comparable approaches on the challenging CALVIN benchmark, testing long-horizon manipulation tasks guided by language instructions. 3) \textbf{Zero-Shot Real-World Transfer:} We achieve successful zero-shot transfer of policies learned entirely offline in the world model with language conditioning to the real environment. \looseness=-1

\section{Related Work}
\label{sec:related_work}

\textbf{Language-Conditioned Imitation Learning for Robotics:} Lynch \etal \cite{lynch2020language} introduce MCIL, which trains a behavior cloning agent using unstructured robot play data. They annotate 1\% of trajectories with human language, making the policy steerable at test-time by natural language commands. Mees \etal \cite{mees2022calvin} introduced the open-source CALVIN environment, a benchmark for evaluating language-conditioned, long-horizon multi-task robotic manipulation. It includes a simulator with \(\sim\)24 hours of teleoperated play data and 20K language task annotations. HULC~\cite{mees2022matters} introduces a framework for hierarchical language-conditioned imitation learning using contrastive learning to ground language instructions to robotic actions, as well as other improvements like using a multimodal transformer encoder (discussed below). HULC outperforms MCIL on the CALVIN benchmark~\cite{zhou2023language}. HULC++~\cite{mees2023grounding} further enhances long-horizon performance by integrating affordance prediction and motion planning into the control loop, only deferring to the learned HULC controller when the end-effector is near the manipulation target. SPIL~\cite{zhou2023language} improves upon the state-of-the-art by labeling all actions in the dataset and probabilistically assigning them to a base skill (translation, rotation, grasping) based on their magnitudes. In this work, we do not incorporate any prior knowledge about the semantic relationship between action dimensions and particular skills, as we aim to maintain a 7-DoF continuous action space. Diffusion Generative Models have gained traction for policy representation in robotics, diffusing actions from Gaussian noise~\cite{chi2023diffusionpolicy,reuss2023goal,xian2023chaineddiffuser}. They can acquire diverse behaviors conditioned on language-goals~\cite{ke20243d,black2023zero,ha2023scaling}. The latest, Multimodal Diffusion Transformer (MDT)~\cite{reuss2024multimodal}, uses Masked Generative Foresight and Contrastive Latent Alignment to boost performance in long-horizon tasks. Although this approach is orthogonal to our work, we acknowledge the potential of diffusion models in both world modeling and behavior learning. In this paper, however, we aim to investigate how to train policies in the latent space of a learned world model.\looseness=-1

\begin{figure*}[t]
	\centering
	\includegraphics[scale=0.72]{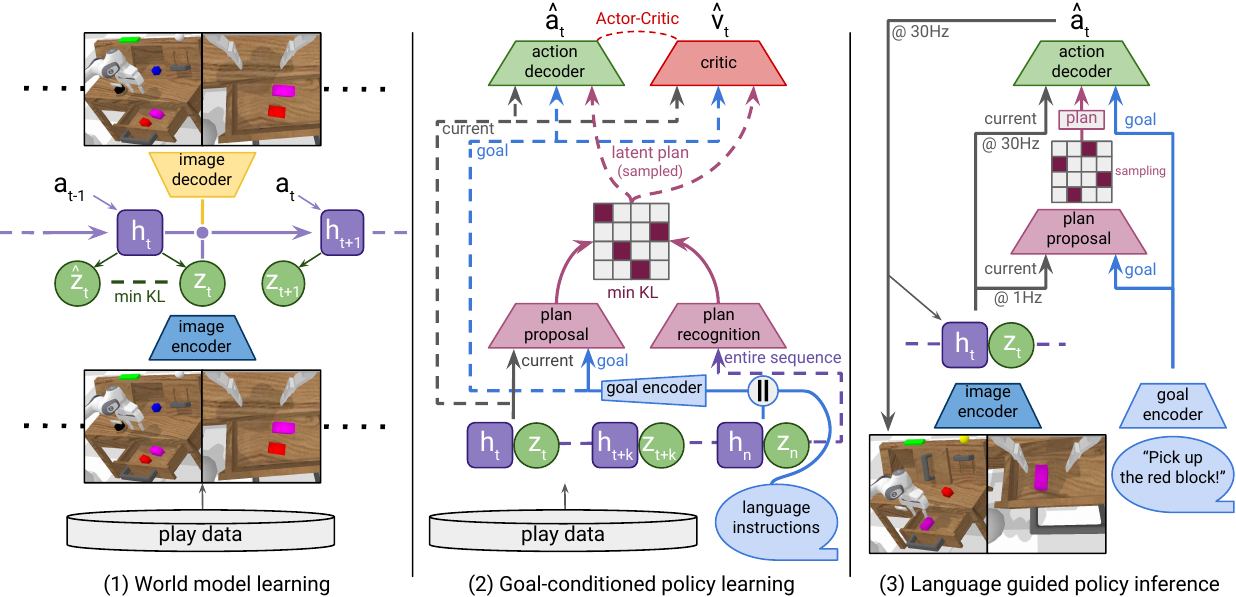}
        \caption{\textbf{\ourmodel} learns a language-guided general-purpose policy within the latent space of a world model. (1) The world model, comprising an image encoder, a Recurrent State-Space Model (RSSM) for dynamics, and an image decoder, transforms play dataset experience into a predictive model that enables behavior learning in the latent state space. (2) The goal-conditioned policy samples latent trajectories and uses \textit{either} a language annotation \textit{or} the final latent state as the goal, with plan recognition and proposal networks being trained to identify and organize behaviors in a latent plan space. The action decoder is intrinsically rewarded by matching the expert's latent trajectory. (3) During inference, the policy acts based on the latent state inferred by the world model from the current observation and is guided by a user's language command.\looseness=-1 }
    \label{fig:model_overview}
    \vspace{-0.5cm}
\end{figure*}

\textbf{Improving Behavior Cloning with Latent Planning:} Lynch \etal \cite{lynch2020learning} introduce GCBC, which augments standard behavior cloning with image-based goal conditioning. GCBC further introduces a planning module which trains a plan-predictor conditioned only on the current and goal states to predict a latent variable produced by a planning module that has privileged access to the entire expert trajectory. At test time, latent variables sampled from the plan predictor are used to condition the policy to achieve improved long-horizon performance. One limitation of this work is the reliance on goal images for conditioning the plan predictor. Zhang \etal \cite{zhang2024language} improve upon the latent planner in GCBC by replacing the discrete latent plan embedding model with a diffusion model. They learn to iteratively denoise the latent plan via a language-conditioned U-Net, achieving superior long-horizon performance over GCBC.\looseness=-1

\textbf{Learning with World Models:} World models have recently emerged as a promising approach to data-driven simulation. The seminal work of Ha and Schmidhuber~\cite{ha2018world} demonstrated the effectiveness of learning world models directly from pixel observations to predict future observations, and then learning a policy inside the latent space of the learned model. Building on this, the Dreamer family of works~\cite{hafner2019dream,hafner2020dreamerv2} introduced a world-model-based RL algorithm that achieved SOTA performance on the Atari benchmark, and recently became the first algorithm to produce diamonds in Minecraft, which requires acting coherently across thousands of time-steps~\cite{hafner2023mastering}. DayDreamer~\cite{wu2023daydreamer} applied the Dreamer algorithm to learning quadrupedal locomotion online on a real robot, demonstrating the robustness of policies trained in the world model to transfer to the real environment. 

Our work builds primarily on DITTO~\cite{demoss2023ditto}, a world-model-based imitation learning algorithm that penalizes an on-policy divergence from expert trajectories in the world model latent space. Specifically, it defines an \textit{intrinsic reward} in the latent space that measures the divergence between agent rollouts and expert demonstrations and then optimizes this intrinsic reward using actor-critic RL. Optimizing this intrinsic reward induces imitation learning that is robust to errors over long-horizon rollouts, effectively mitigating covariate shifts.\looseness=-1

\section{Problem Formulation}
\label{sec:problem}

We investigate goal-conditioned imitation learning in a partially observable Markov decision process (POMDP) with continuous actions and high-dimensional observations from an unknown environment. We model the interaction between the environment and the goal-conditioned policy using a goal-augmented POMDP $\mathcal{M} = (\mathcal{S}, \mathcal{A}, \mathcal{R}, \mathcal{T}, \mathcal{G}, \gamma)$, where $\mathcal{S}$ is the state-observation space, $\mathcal{A}$ is the action space, $\mathcal{R}(s, a)$ is the reward function, $\mathcal{T}(s'|s, a)$ defines state-transition dynamics, $\mathcal{G}$ is the goal space, and $\gamma \in (0, 1)$ is the discount factor. The agent receives visual observations instead of direct state access and aims to maximize the expected discounted sum of extrinsic rewards $\mathbb{E}[\sum_t \gamma^t r_t]$, which it cannot directly observe. We conduct offline learning using a large, unlabeled, and undirected, fixed play dataset $\mathcal{D} = \{ (s_1, a_1), \ldots, (s_T, a_T)\}$. This dataset is relabeled into long temporal state-action streams~\cite{andrychowicz2017hindsight}. Each visited state is treated as a ``reached goal state'', resulting in a trajectory dataset $\mathcal{D} = \{ \tau_i = (s_t, a_t)_{t=0}^k \}_{i=1}^N$. This transformation creates $\mathcal{D}_{\text{play}} = \{ (\tau, s_g )_i \}_{i=0}^{|\mathcal{D}_{\text{play}}|}$, pairing each goal state $s_g$ with the corresponding demonstration trajectory $\tau$.
We adopt the multi-context imitation learning approach from play data by Lynch \etal \cite{lynch2020language} to utilize language annotations. This method shows that combining a few random windows with language-based retrospective instructions helps learn a unified, language-conditioned visuomotor policy for various robotic tasks. Free-form language instructions $l \in \mathcal{L}$ guide the policy, providing flexible and natural task descriptions.\looseness=-1

\section{\ourmodel}
\label{sec:approach}

In this section, we introduce \ourmodel. Training consists of two phases: first, a world model is learned from the unlabeled play dataset $\mathcal{D}$. Next, an actor-critic agent is trained within this learned world model to acquire a goal-conditioned policy by guiding imagined sequences of latent model states to match the latent trajectory of the expert demonstrations. During inference, the language-conditioned policy $\pi_{\theta} (a_t \mid s_t, l)$, trained entirely in the latent space of the world model, successfully transfers to the real environment. \figref{fig:model_overview} shows an overview of the approach.\looseness=-1 

\subsection{World Model Learning}
World models capture an agent's experience into a predictive model, enabling behavior learning without direct interaction with the environment. By converting high-dimensional images into compact state representations, they allow efficient, parallel long-term predictions in latent space. While our method is agnostic regarding the choice of the world-model architecture, we follow DITTO by adopting the DreamerV2 architecture~\cite{hafner2020dreamerv2} as the backbone for \ourmodel, which includes an image encoder, a recurrent state-space model (RSSM) for learning transition dynamics, and a decoder for reconstructing observations from latent states. To learn robotic manipulation skills, we use observations from static and gripper-mounted cameras, with separate CNN encoders and transposed CNN decoders for each. These encodings are concatenated before being passed to the RSSM, which generates a sequence of deterministic recurrent states $(h_t)_{t=0}^T$. Each state defines two distributions over stochastic hidden states: the stochastic posterior state $z_t$, determined by the current observation $x_t$ and recurrent state $h_t$, and the stochastic prior state $\hat{z}_t$, trained to approximate the posterior without the current observation. By predicting $\hat{z}_t$, the model learns the environment dynamics. The combined model state, comprising deterministic and stochastic components $\hat{s}_t = (h_t, z_t)$, reconstructs the observation. The RSSM includes the following modules:\looseness=-1
\begin{equation}\label{modelcomponents}
\hspace*{-0.2cm} 
\text { RSSM } \begin{cases}\text{Recurrent state:}     &h_t     = f_\phi(\hat{s}_{t-1}, a_{t-1}) \\ \text{Representation model:}  &z_t     \sim q_\phi(z_t \mid h_t, x_t) \\ \text{Dynamics predictor:}  &\hat{z}_t \sim p_\phi(\hat{z}_t \mid h_t)\\  \text{Image decoder:}   &\hat{x}_t \sim p_\phi(\hat{x}_t \mid \hat{s}_t)\end{cases}
\end{equation}

The prior and posterior models predict categorical distributions, optimized using straight-through gradient estimation~\cite{bengio2013estimating}. All modules are implemented as neural networks parameterized by $\phi$, and are optimized jointly by minimizing the negative variational lower bound~\cite{kingma2013auto}:\looseness=-1
\begin{equation}\label{ELBO}
\begin{split}
     \min_{\phi} \mathbb{E}_{q_\phi(z_{1:T} \mid a_{1:T}, x_{1:T})} \Bigg[
     \sum_{t=1}^T -\log p_\phi(x_t \mid \hat{s}_t&) \\ 
     + \beta \kl{q_\phi(z_t \mid \hat{s}_t)}{ p_\phi(\hat{z}_t \mid h_t)}
    &\Bigg].
\end{split}
\end{equation}

After training, the world model can operate without input observations by using the prior $\hat{z}$ instead of the posterior $z$. This allows the model to generate unlimited imagined trajectories of the form $\{(h_t, \hat{z}_t, a_t)_{t=0}^H \}$, where $H$ is the time horizon for imagination.\looseness=-1
\subsection{Behavior Learning}
\ourmodel~employs an actor-critic agent to learn long-horizon, language-conditioned behaviors within the latent space of its world model. It learns a goal-conditioned policy $\pi_{\theta}(a_t \mid s_t, g)$ and a value function $v_\psi(s_t, g)$ from the dataset $\mathcal{D}_{\text{play}}$, where $s_t$ denotes the current latent state and $g \in \mathcal{G}$ represents a free-form language instruction $l$ \textit{or} a latent goal state $s_g$. In this section, we outline the design choices of our model, explain how to intrinsically reward the agent for matching expert latent state-action pairs over a trajectory, and detail the optimization process for the actor and critic.\looseness=-1

\textbf{Observation and Action Spaces} \hspace{0.1cm} Our goal-reaching policy builds on the HULC architecture by Mees \etal \cite{mees2022matters}, enhancing language-conditioned imitation learning from unstructured data. Unlike HULC, which uses high-dimensional image observations, our method utilizes the latent state space of the world model for its compact representations. In addition to encoding RGB images from static and gripper cameras, \ourmodel~encodes compact states of latent trajectories from the current to the goal state. The action space is a 7-DoF continuous space, including relative XYZ positions, Euler angles, and a gripper action.\looseness=-1

\textbf{Latent Plan Encoding} \hspace{0.1cm} Learning control with free-form imitation data faces the challenge of multiple valid trajectories connecting the same $(s_t, s_g)$ pairs. To address this, we use a sequence-to-sequence conditional variational auto-encoder (seq2seq CVAE)~\cite{lynch2020learning} to encode contextual latent trajectories into a latent ``plan'' space. This space is determined by two stochastic encoders: one for recognizing plans during training and the other for proposing plans during inference. The plan recognition encoder identifies the performed behavior from the entire sequence, while the plan proposal encoder generates possible behaviors from the initial and final states (see the second column of \figref{fig:model_overview}). Minimizing the KL divergence between these encoders, referred to as $\mathcal{L}_{\text{KL}}$, ensures the plan proposal encoder accurately reflects observed behaviors. Similar to HULC, we use a multimodal transformer encoder to develop a contextualized representation of latent trajectories, mapping them into a vector of multiple latent categorical variables, optimized with straight-through gradients.\looseness=-1

\textbf{Semantic Alignment of Latent Trajectories and Language} \hspace{0.1cm} Addressing the symbol grounding problem~\cite{harnad1990symbol} requires relating language instructions to the robot's perception and actions. Aligning instructions with visual observations is crucial for distinguishing similar objects like colored blocks. Unlike HULC, which aligns visual features with language features, we align the world model's latent features with language features. This involves maximizing cosine similarity between latent and language features while minimizing similarity with unrelated instructions, using a contrastive loss similar to CLIP~\cite{radford2021learning} and incorporating in-batch negatives for efficient training. We denote this objective as $\mathcal{L}_{\text{contrast}}$.\looseness=-1

\textbf{Reward} \hspace{0.1cm} Unlike behavior cloning approaches, which rely exclusively on supervised action labels for training, \ourmodel~leverages a simple intrinsic reward defined within the latent space of the world model to address the problem of covariate shift. To align the agent's behavior with the expert, we employ a reward function proposed by DeMoss \etal \cite{demoss2023ditto} that encourages the agent to match the expert's state-action pairs in the latent space. This intrinsic reward is defined as a modified inner product on the latent state representations, encouraging similarity without requiring exact matches. This lets the agent explore different trajectories in the world-model latent space, while aligning its behavior to the demonstrator over the entire training horizon. The reward at each step $t$ is given by:
\begin{equation}\label{intreward}
    r^{int}_t(s^E_t, s^\pi_t) = \frac{s^E_t \cdot s^\pi_t}{\max(\|s^E_t\|, \|s^\pi_t\|)^2} 
\end{equation}

\textbf{Action and value models} \hspace{0.1cm} Our stochastic actor, parameterized by $\theta$, samples actions based on the current latent state $s_t$ and a latent goal $g$, aiming to maximize expected rewards (Eqn.~\ref{intreward}). The deterministic critic, parameterized by $\psi$, predicts the expected discounted future rewards, providing value estimates for the same latent state and goal. The models are defined as:\looseness=-1
\begin{equation}
\begin{aligned}
    &\text{Actor:} \quad a_t \sim \pi_\theta(a_t \mid s_t, g), \quad \\ 
    &\text{Critic:} \quad v_\psi(s_t, g) \approx \mathbb{E}_{\pi_\theta, p_\phi} \left[ \sum_{t=0}^H \gamma^t r^{int}_t \right].
\end{aligned}
\end{equation}
Fully-connected neural networks are used for the actor and critic. The action model outputs a tanh-transformed Gaussian~\cite{haarnoja2018sac}, facilitating reparameterized sampling for backpropagation.
The world model is fixed during behavior learning, ensuring that the agent gradients do not alter its representations.\looseness=-1
 \setlength{\tabcolsep}{4pt}
 \begin{table*}[b]
 \vspace{-0.4cm}
  \centering
  \small
  \begin{tabular}{|p{2.1cm} |c| c| c| c| c|  c| }
  \hline
  Method  & \multicolumn{6}{|c|}{No. Instructions in a Row (1000  chains)}\\
  \cline{2-7}
&          1 & 2 & 3 & 4 & 5 & Avg. Len.\\
  \hline
 GCBC~\cite{lynch2020learning}     & 56.7\% (3.4)& 20.9\% (1.9)& 7.3\% (2.9)& 1.6\% (0.5)& 0.04\% (0.0)& 0.63 (0.4)\\
 MCIL~\cite{lynch2020language}     & 70.3\% (3.1)& 40.2\% (2.9)& 21.6\% (4.1)& 11.1\% (1.9)& 5.4\% (1.0)& 1.48  (0.3)\\
 HULC~\cite{mees2022matters}     & 77.6\% (1.1) & 58.05\% (0.3) & 41.6\% (1.3) & 29.8\% (0.9) & 20.0\% (1.5)& 2.27 (0.05)\\
 \ourmodel~(ours)            & \textbf{80.7}\% (1.0) & \textbf{59.3}\%  (1.4)& \textbf{42.6}\%  (1.8)& \textbf{30.7}\%  (1.2)& \textbf{21.1}\%  (0.8)& \textbf{2.34}  (0.05)\\
  \hline
No DITTO     &  71.8\% (1.3)& 45.6\% (1.5)& 27.8\% (1.9)& 14.2\% (1.8)& 8.7\% (0.7)& 1.68 (0.02)\\
No latent plan     & 76.5\% (2.5)& 50.5\% (0.9)& 28.7\% (1.3)& 15.9\% (0.8)& 11.0\% (0.7)& 1.81 (0.01)\\
No alignment    & 73.3\% (1.5)& 52.6\% (0.9)& 39.8\% (2.0)& 27.3\% (1.3)& 17.2\% (0.9)& 2.05 (0.06)\\

 \hline
  \end{tabular}
  \caption{Performance of all models on environment \textit{D} of the CALVIN Challenge with an ablation study of key components evaluated over three seeds. All models receive $64\times64$ resolution RGB images from both a static and gripper camera. Standard deviation in parentheses.
}
  \label{tab:roboexp}
\end{table*}

\textbf{Learning Objective} \hspace{0.1cm} The objective in training the critic is to regress the computed value estimate to the $\lambda$-target~\cite{suttonbarto}, denoted $V^\lambda$ (refer to~\cite{hafner2020dreamerv2} for the definition):\looseness=-1
\begin{equation}\label{critictarget}
    \mathcal{L}(\psi) \, \dot{=} \,
    \mathbb{E}_{\pi_\theta, p_\phi}\left[ \sum_{t=1}^{H-1}\tfrac{1}{2} ( v_\psi(\hat{s}_t) - sg(V^\lambda_t))^2\right]
\end{equation}
where $sg(\cdot)$ is the stop-gradient operator.  Moreover, a target network, which updates its weights slowly, is used to provide stable value bootstrap targets~\cite{mnih2015human}. The actor's objective function uses backpropagation through the learned dynamics to maximize value estimates while minimizing the KL divergence between plan encoders and the contrastive loss for aligning latent trajectories with language. For sampled windows without language annotations, the contrastive loss is omitted:\looseness=-1
\begin{equation}\label{eq:action_objective}
\hspace*{-0.2cm} \mathcal{L}(\theta) \, \dot{=} \, \mathbb{E}_{\pi_\theta, p_\phi} \left[ \sum_{\tau=t}^{t+H} \left( - V_{\lambda}(s_\tau) + \alpha_1\mathcal{L}_{KL} + \alpha_2 \mathcal{L}_{\text{contrast}} \right) \right]
\end{equation}

\section{Experimental Results}
\label{sec:result}
We evaluate \ourmodel~in learning language-conditioned imitation learning in both simulated and real-world settings. The objectives of our experiments are threefold: (i) to determine the model's performance in executing long-horizon language-specified tasks, (ii) to ablate components of the objective function and analyze their impact, and (iii) to assess its scalability to real-world robotics tasks.\looseness=-1
\subsection{Experiments in Simulation}
We conduct our simulation experiments in environment \textit{D}  of the CALVIN Challenge~\cite{mees2022calvin}, which offers six hours of unstructured tele-operated play data to train a 7-DoF Franka Emika Panda robot arm. CALVIN features 34 distinct subtasks and evaluates 1000 unique instruction chain sequences. The agent's objective is to sequentially solve up to five language instructions using only onboard sensors, requiring it to transition between subgoals based solely on language cues. Notably, only 1\% of this dataset is annotated with language instructions through crowd-sourcing.

\textbf{Evaluation Protocol} \hspace{0.1cm} We compare our model with the following language-conditioned robotic imitation learning approaches over unstructured data: \textbf{GCBC}~\cite{lynch2020learning}: 
Goal-conditioned behavior cloning does not model latent plans but instead maximizes the likelihood of each action as reconstruction loss at each time step. \textbf{MCIL}~\cite{lynch2020language}: Multi-context imitation learning employs a sequential CVAE to predict the next actions from image or language-based goals by modeling reusable latent sequences of states as latent plans. \textbf{HULC}~\cite{mees2022matters}: Hierarchical universal language conditioned policy provides a strong baseline by building upon MCIL, introducing several improvements, such as adding a self-supervised contrastive loss for alignment of video and language representation. Our approach differs from the above models by not regressing action labels. Instead, we learn a world model from the data, before training an agent to match the expert's latent trajectory in the world model. To reduce the computational burden of \ourmodel' world model, we standardize the input resolution of all images to \(64 \times 64\). For fair comparison, all baselines are trained at this resolution, which sometimes lowers the original implementations' resolutions (e.g., \(200 \times 200\) for static and \(84 \times 84\) for the gripper camera).\looseness=-1

\tabref{tab:roboexp} contains quantitative results of the average success rate of each model over 1,000 instruction chains across three seeded runs.
\begin{figure*}[t]
\centering	\includegraphics[scale=0.72]{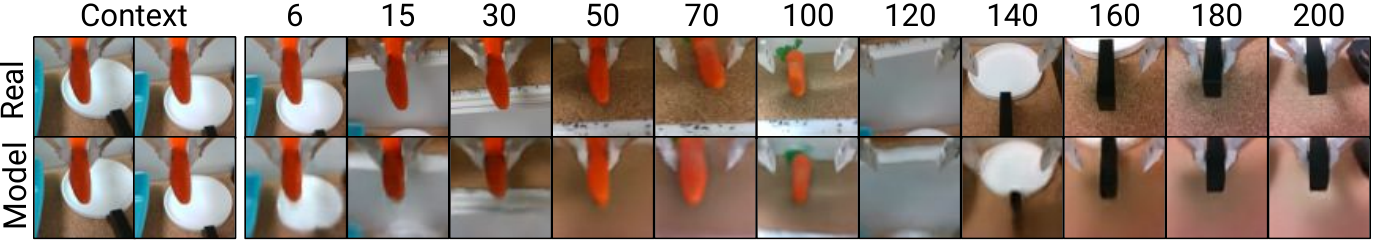}
\caption{\textbf{World Model Long-Horizon Predictions.} Using the first five images of a hold-out trajectory as context, the world model predicts the next 195 steps using its latent dynamics, given only the actions. Trained on sequences of horizon 50, our model makes precise long-term predictions, aiding efficient behavior learning in a compact latent space. Only gripper camera reconstructions are visualized.}
\vspace{-0.5cm}
\label{fig:wm_real}
\end{figure*}
\ourmodel~successfully performs multi-stage, long-horizon, language-conditioned robotic manipulation tasks, and outperforms all the baselines. This experiment shows that our agent can optimize its intrinsic reward, which measures its deviation from expert latent trajectories. This capability allows the agent to recover from its mistakes over multiple time steps and successfully chain more tasks together in sequence. 

In our first ablation, we examine the impact of our latent matching reward function. Specifically, we supervise our agent using behavior cloning with mean squared error instead of optimizing for intrinsic rewards. This ablation demonstrates a significant drop in our model's performance under these conditions and indicates that our reward formulation effectively reduces covariate shift. For the second ablation study, we explore the effect of learning a latent plan space on long-horizon behavior. For this test, we removed the plan proposal and recognition components from the actor's architecture. The results show that, although the agent starts the trials relatively well, it struggles to complete longer-horizon tasks. Lastly, in our final ablation, we omit the alignment objective between language and latent state features. While this variant maintains some long-horizon performance, it performs slightly worse at each step compared to our full model. This is evident in its occasional manipulation of blocks of the wrong color due to a suboptimal association between instructions and the scene.\looseness=-1
\begin{figure}[b]
\vspace{-0.5cm}
\centering	\includegraphics[width=1.0\columnwidth]{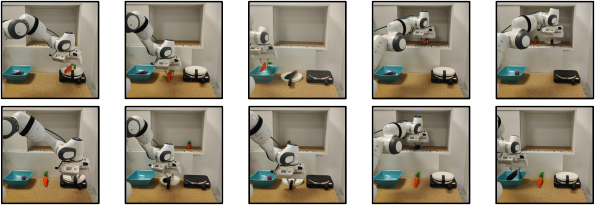}
\caption{\textbf{Real-world Tasks.} Examples from left to right are: placing the carrot onto the pan, lifting the carrot from the table, dropping the carrot into the bowl, storing the carrot in the cabinet, placing the pan into the cabinet, setting the pan onto the stove, adding the eggplant to the pan, picking up the pan from the table, taking the eggplant from the cabinet, and putting the eggplant into the bowl.\looseness=-1 }
\label{fig:real_skills}
\end{figure}

\subsection{Experiments in Real-World}

To assess the performance of \ourmodel{} at handling various real-world tasks, we conduct experiments using a Franka Emika Panda robot arm within a 3D tabletop environment. This environment includes a table with a stove, a bowl, a cabinet, a carrot, and an eggplant. We collected a play dataset by recording three hours of teleoperated interactions with the robot arm, using a VR controller to guide its movements. During recording, we captured RGB observations from both static and gripper cameras. We used language annotation for less than 1\% of the data (\(\sim\)2800 random windows). Annotators described behaviors in randomly sampled portions of the interaction data. This dataset encompasses over 22 unique tasks for manipulating the available objects.\looseness=-1

First, we used the offline play dataset to learn a generative world model. As shown in \figref{fig:wm_real} our model demonstrates competence in long-horizon predictions. To assess its long-term prediction abilities, we applied the representation model to the first five images of hold-out trajectories to establish context. Given an action trajectory, we then tasked the model with predicting forward for 195 steps using its latent dynamics, despite being trained only with a horizon length of 50. The accuracy of the reconstructed images decoded from the latent trajectory of our recurrent state-space model indicates that our world model is well-suited for learning behaviors within its compact latent space.\looseness=-1        

We trained our language-conditioned policy within the latent space of the world model and subsequently assessed each task through~20 rollouts, comparing its performance to HULC~\cite{mees2022matters} under the same conditions. We utilized various neutral starting positions to avoid biasing the policy toward any specific task. This neutral setup eliminates the correlation between the robot’s initial state and the queried task and ensures that the agent depends entirely on language cues to comprehend and execute the tasks. The success rates are reported in~\tabref{tab:rwresults}. For brevity, we combined the success rates for tasks involving either the carrot or eggplant under the category ``vegetable''. However, each task was evaluated individually (i.e., if the robot picks up the carrot when the instruction was to pick up the eggplant, it does not count as a success). Moreover, we assessed our model and HULC on 20 unique chains of 5 instructions in a row to evaluate their multi-stage long-horizon capabilities. We reported the average number of successfully completed sequential tasks, as shown in~\tabref{tab:rwresults}. Our findings indicate that \ourmodel~outperforms HULC in both individual and multi-stage long-horizon tasks, demonstrating superior performance in recovering from its own mistakes across multiple time steps and reducing covariate shift.

Finally, we analyzed LUMOS's predicted action trajectories by simulating their outcomes in the learned world model. The performance gap between real and simulated environments stems from inaccuracies in the world model, particularly for tasks with lower real-world success rates. We believe that training the world model with more comprehensive data could improve its robustness and reduce this gap. \looseness=-1

 \begin{table}[t]
      \centering
      \scriptsize
\begin{tabular}{p{3.5cm} | >{\centering\arraybackslash}p{1.5cm} | >{\centering\arraybackslash}p{1.1cm} | >{\centering\arraybackslash}p{1.4cm}}
            \toprule
            Task / Method & \ourmodel~(ours)  & HULC~\cite{mees2022matters}& World Model\\
            \midrule
            Lift the vegetable from the table   &  67.5\% & 62.5\% & 72.5\%
            \\
            Lift the vegetable from the pan   &  75\% & 72.5\% & 77.5\%
            \\
            Lift the vegetable from the bowl   &  72.5\% & 67.5\% & 75\%
            \\
            Lift the vegetable from the cabinet   &  55\% & 50\% & 70\%
            \\
            Lift the pan from the table   &  65\% & 75\% & 70\%
            \\
            Lift the pan from the stove   &  80\% & 70\% & 75\%
            \\
            Lift the pan from the cabinet   &  35\% & 35\% & 60\%
            \\
            Place the vegetable onto the table   &  90\% &  85\% & 92.5\%
            \\
            Place the vegetable onto the pan   &  62.5\% &  65\% & 77.5\%
            \\
            Place the vegetable into the bowl   &  85\% &  75\% & 80\%
            \\
            Place the vegetable into the cabinet   &  70\% &  60\% & 77.5\%
            \\
            Place the pan onto the table  &  70\% &  65\% & 80\%
            \\
            Place the pan onto the stove  &  65\% &  55\% & 85\%
            \\
            Place the pan into the cabinet  &  55\% &  50\% & 70\%
            \\
            \midrule
            Average over tasks &  \textbf{67.68}\% & 63.39\% & 75.89\%
            \\
            \midrule
            Average no. of sequential tasks & \textbf{2.05}& 1.90 & -
            \\
            \bottomrule
          \end{tabular}
          \caption{The average success rate of language-conditioned general-purpose policies in the real world setup.}  
        \label{tab:rwresults}
        \vspace{-3em}
   \end{table}

\section{Conclusion and Limitations}
\label{sec:conclusion}
In this paper, we propose \emph{Language-Conditioned Imitation Learning via World Models} (\ourmodel), which leverages a learned world model's latent space for fully offline, multi-step on-policy learning. Practicing skills in long-horizon model-based rollouts allows our model to recover expert performance without any online interaction. Our approach demonstrates superior performance on the CALVIN benchmark, achieving robust long-horizon manipulation guided solely by language instructions. \ourmodel~serves as a proof of concept, demonstrating that dynamics and policies learned offline within a world model can successfully transfer zero-shot to real environments.\looseness=-1

Although \ourmodel~performs well, it has several aspects that warrant future research. One difficulty with world-model-based approaches is that it is hard to know a priori whether the world model's representations and dynamics are of sufficient quality to produce strong policies. Additionally, it has been observed that increasing image input resolution substantially improves the final performance. However, training world models at higher resolution significantly increases the compute resources required to run experiments. \looseness=-1


\newpage

\footnotesize

\newpage
\section{Appendix}
\subsection{Hyperparameters and Training Details of the World Model}

To develop the world model of \ourmodel, we used DreamerV2~\cite{hafner2020dreamerv2} as our foundation. While DreamerV2 models Atari game environments, our world model is specifically tailored for robotic manipulation. As a result, \ourmodel~incorporates distinct encoders and decoders for the static and gripper cameras. We concatenate their encodings and employ a fully-connected layer to fuse them before passing the features to the RSSM. This results in our model comprising approximately 40 million parameters.
As described in (Eqn. 2), all components of the world model are trained jointly using a modified ELBO objective. Following~\cite{hafner2020dreamerv2}, we incorporate KL balancing in this training process, which is used to control the regularization of the prior and posterior relative to each other using a parameter $\delta$:
\begin{equation}\label{kl_balancing}
    \kl{q}{p} = \delta \underbrace{\kl{q}{sg(p)}}_{\text{posterior regularizer}} + (1-\delta)\underbrace{\kl{sg(q)}{p}}_{\text{prior regularizer}}
\end{equation}
where $sg(\cdot)$ is the stop-gradient operator. KL balancing is necessary because the prior and posterior should not be regularized at the same rate; the prior must update more quickly towards the posterior, which encodes more information. This approach ensures the prior better approximates the aggregate posterior, enhancing the model's performance. 

\begin{table}[b]
\centering
\begin{tabular}{lcc}
\toprule
\textbf{Name} & \textbf{Symbol} & \textbf{Value} \\
\midrule
Batch size & $B$ & 50 \\
Sequence length & $L$ & 50 \\
Deterministic latent state dimensions & --- & 1024 \\
Discrete latent state dimensions & --- & 32 \\
Discrete latent state classes & --- & 32 \\
Latent dimensions & $k$ & 2048 \\
KL loss scale & $\beta$ & 0.3 \\
KL balancing & $\delta$ & 0.8 \\
RSSM reset probability & $\zeta$ & $0.01$ \\
World model learning rate & --- & $3\cdot10^{-4}$ \\
Gradient clipping & --- & 100 \\
Adam epsilon & $\epsilon$ & $10^{-5}$ \\
Weight decay (decoupled) & --- & $5\cdot10^{-2}$ \\
\bottomrule
\end{tabular}
\vspace{0.5em}
\caption{hyperparameters for learning the world model.}
\label{tab:wm_hparams}
\end{table}

To achieve uniform coverage of various segments of play data while modeling the world, we randomly sample the start index of each training sequence within the episode and clip it to prevent exceeding the episode length minus the training sequence length.

Since our approach focuses on learning from offline, unstructured teleoperated play data, it includes very long-horizon episodes and consequently very few reset states. To address this, we randomly reset the hidden state of the RSSM during training with a probability $\zeta$. This ensures that the world model learns to attend to both the current observation and the trajectory's history while learning from long-horizon streams of data.

In all simulated and real-world experiments, we use similar hyperparameters to learn the world model, as outlined in~\tabref{tab:wm_hparams}.

The latent state consists of deterministic and stochastic components, $\hat{s}_t = (h_t, z_t)$. Flattening the stochastic sample from 32 categorical distributions, each with 32 classes, results in a sparse binary vector of length 1024. Consequently, the latent size for each time step becomes $k=2048$.

\subsection{Hyperparameters and Training Details of the Agent}
Since our goal-reaching policy is inspired by the HULC architecture~\cite{mees2022matters}, we maintain similar design choices wherever possible for a fair comparison. Any deviations from the original design are specified in this section. Notably, HULC operates on raw observations, whereas our model first infers the latent state trajectories by the learned world model and then takes latent trajectories as input.

\textbf{Perceptual Encoder} \hspace{0.1cm} 
While HULC encodes the sequence of visual observations from both static and gripper cameras \( X_{\{static, gripper\}} \in \mathbb{R}^{T \times H \times W \times 3} \) using separate convolutional encoders, \ourmodel~first leverages its learned world model to infer the latent state trajectories from the camera observations and then encodes the sequence of latent states \( S \in \mathbb{R}^{T \times k} \) using a fully connected encoder with two layers, featuring a hidden size of 1024 and an output size of 128 units with ReLU activations. Both methods map to the same output space, forming a perceptual representation \( V \in \mathbb{R}^{T \times d} \), where \( T \) represents the sequence length and \( d \) the feature dimension.

\textbf{Goal Encoder} \hspace{0.1cm} The goal-conditioned policy samples latent trajectories and uses \textit{either} a language annotation \textit{or} the final latent state as the goal. For each goal type, we employ a goal encoder similar to HULC. Notably, to transform raw text into a semantically meaningful vector space, we use the paraphrase-MiniLM-L3-v2 model~\cite{reimers-2019-sentence-bert}. This model is trained on paraphrase datasets primarily sourced from Wikipedia. It supports a vocabulary of $30,522$ words and converts any length of sentence into a $384$-dimensional vector. Both goal encoders map their respective inputs to a latent goal representation in \( \mathbb{R}^{g} \), where \( g \) is the latent goal dimension.

\textbf{Plan Recognition and Proposal} \hspace{0.1cm} While MCIL~\cite{lynch2020language} proposed using bidirectional recurrent neural networks (RNNs) as the plan recognition network to encode randomly sampled play sequences into a latent Gaussian distribution, HULC demonstrated improved performance with a multimodal transformer encoder~\cite{vaswani2017attention}, incorporating positional embeddings to capture temporal information, and representing latent plans as vectors of multiple categorical variables~\cite{hafner2020dreamerv2}. We adopt HULC's method by passing the encoded perceptual representation \( V \) to the transformer, leveraging its scaled dot-product attention mechanism, which allows elements within a sequence to attend to each other. This transformer encoder consists of two blocks, eight self-attention heads, and a hidden size of $2048$ units. The plan proposal network is implemented using four fully connected layers, each with 2048 units and ReLU activations. To minimize the KL divergence between these encoders, referred to as $\mathcal{L}_{\text{KL}}$, we employ the KL balancing technique (Eqn.~\ref{kl_balancing}) again, ensuring that the KL loss decreases faster with respect to the prior than the posterior.
\begin{table}[b]
\centering
\begin{tabular}{lcc}
\toprule
\textbf{Name} & \textbf{Symbol} & \textbf{Value} \\
\midrule
Batch size & $B$ & 512 \\
Sequence length & $T$ & 32 \\
Minimum window length & --- & 20 \\
Maximum window length & --- & 32 \\
Feature dimensions & $d$ & 128 \\
Latent goal dimensions & $g$ & 32 \\
Discrete latent plan dimensions & --- & 32 \\
Discrete latent plan classes & --- & 32 \\
KL loss scale & $\alpha_1$ & 0.1 \\
Contrastive loss scale & $\alpha_2$ & 3.0 \\
KL balancing & $\delta$ & 0.8 \\
Discount factor  &$\gamma$  &$0.995$\\
$TD(\lambda)$ parameter         &$\lambda$  &$0.95$\\
Optimizer & --- & Adam \\
Actor learning rate & --- & $2\cdot10^{-4}$ \\
Critic learning rate & --- & $3\cdot10^{-4}$ \\
Slow critic update interval & --- & $100$ \\
\bottomrule
\end{tabular}
\vspace{0.5em}
\caption{hyperparameters for training the actor and critic of \ourmodel.}
\vspace{-1.5em}
\label{tab:agent_hparams}
\end{table}

\textbf{Alignment of Latent Trajectories and Language} \hspace{0.1cm} \ourmodel~maximizes the cosine similarity between the latent trajectory features of sequence \(i\) and its corresponding language features while minimizing the similarity with unrelated language instructions within the same batch. This is achieved using a contrastive loss similar to CLIP~\cite{radford2021learning}. First, the latent trajectory features ($x_i$) and language features ($y_j$) are normalized, and the cosine similarity between each pair is computed. The logits for each batch form an \(M \times M\) matrix, where each entry is defined as:
\[
\text{logit}(x_i, y_j) = \text{cos\_sim}(x_i, y_j) \cdot \exp(\tau), \quad \forall (i,j) \in \{1,2,\ldots,M\},
\]
with \(\tau\) being a trainable temperature parameter. Only the diagonal entries of this matrix, which represent true pairs, are considered positive examples. The contrastive loss is computed as follows:
\begin{align}
    \mathcal{L}_{\text{contrast}} &= \frac{1}{2} \Bigg( 
        \sum_{i=1}^{M} \text{CrossEntropy}(\textit{logits\_per\_latent}(i), i) \notag \\
        &\quad + \sum_{i=1}^{M} \text{CrossEntropy}(\textit{logits\_per\_text}(i), i) 
    \Bigg)
\end{align}

where \(\textit{logits\_per\_latent}\) represents the matrix of logits with rows corresponding to latent features and columns to language features, and \(\textit{logits\_per\_text}\) is the transpose of this matrix.

\textbf{Action Decoder} \hspace{0.1cm} The actor decodes the actions to be executed at each time step, conditioned on the current and goal latent states as well as the sampled latent plan. These three inputs are concatenated and passed as input to the action decoder. While HULC parameterizes the action decoder as a recurrent neural network (RNN) that outputs parameters of a discretized logistic mixture distribution~\cite{salimans2017pixelcnn++}, our action decoder is significantly different: it is not recurrent and it outputs parameters of a tanh-transformed Gaussian distribution~\cite{haarnoja2018sac}. HULC's action decoder consists of $2$ RNN layers, each with a hidden dimension of $2048$, utilizing $10$ distributions to predict $10$ classes per dimension. This design results in their action decoder having approximately $15$ million parameters. In contrast, our action decoder is composed of eight fully connected layers, each with a hidden size of $256$ units. This architecture results in a more compact model with approximately $1$ million parameters. Unlike HULC, which needs a recurrent network to capture temporal patterns due to its reliance on action labels at each time step, our action decoder leverages the learned world model that encapsulates the temporal structure of latent state-action trajectories. This allows \ourmodel~to maximize intrinsic rewards and recover from mistakes across multiple time steps by matching the expert latent trajectory, thereby eliminating the need for a recurrent architecture.

\textbf{Critic} \hspace{0.1cm} The critic of \ourmodel~consists of eight fully connected layers, each with a hidden size of $1024$ units. It predicts the discounted sum of future rewards (state value) achieved by the actor, given the current and goal latent states. As detailed in Section 4, we use temporal-difference (TD) learning~\cite{sutton1988learning}, training the critic towards a value target constructed from intermediate rewards and future state predictions. While 1-step targets are common, our model leverages the ability to generate multi-step on-policy trajectories, allowing us to use n-step targets for faster reward incorporation. This is achieved using the $\lambda$-target~\cite{suttonbarto}:

\begin{equation*}\label{lambdareturn}
   V^{\lambda}_t = r_t + \gamma \left( (1-\lambda)v_\psi(\hat{s}_{t+1}) + \lambda V^\lambda_{t+1}\right), \quad V^\lambda_{t+H} = v_\psi(\hat{s}_{t+H}).
\end{equation*}

The hyperparameter \(\lambda\) determines the horizon for TD learning. When \(\lambda = 0\), it results in 1-step TD learning. A value of \(\lambda = 1\) corresponds to unbiased Monte Carlo returns. Intermediate values of \(\lambda\) provide an exponentially weighted sum of n-step returns. We set \(\lambda = 0.95\) to prioritize long-horizon targets over short-horizon targets.

As reported in~\tabref{tab:agent_hparams}, during training, we randomly sample windows with lengths between 20 and 32 and pad them to a maximum length of 32 to set the training horizon for the agent.

\begin{figure}[t]
\centering	\includegraphics[width=0.9\columnwidth]{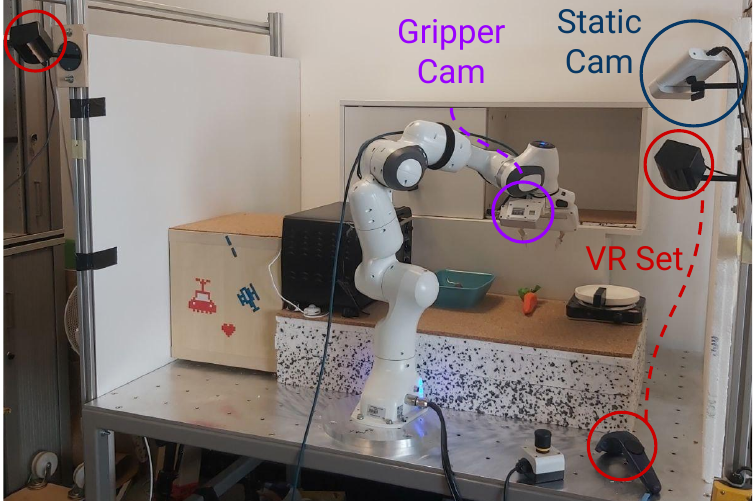}
\caption{Visualization of the complete real-world robot setup. This figure highlights the static and gripper cameras, as well as the VR controller and tracking system.}
\vspace{-1.0em}
\label{fig:rw_setup}
\end{figure}
\subsection{Modifications to Baselines for Fair Comparisons} To efficiently handle the computational demands associated with increasing the resolution of LUMOS' world model, we standardize the input resolution of all images to \(64 \times 64\). Consequently, all baselines are trained at this same resolution for a fair comparison. This reduces the resolution from the original implementations (e.g., \(200 \times 200\) for the static camera and \(84 \times 84\) for the gripper camera). To adapt the vision networks of our baselines to \(64 \times 64\) images, we made the following changes to their convolutional layers:
\begin{itemize}
  \item The first convolutional layer was changed from a \(8 \times 8\) kernel with stride 4 to a \(4 \times 4\) kernel with stride 2.
  \item The second convolutional layer was adjusted from a \(4 \times 4\) kernel with stride 2 to a \(2 \times 2\) kernel with stride 2.
  \item The third convolutional layer was modified from a \(3 \times 3\) kernel with stride 1 to a \(2 \times 2\) kernel with stride 2.
\end{itemize}
While our model does not use any image augmentation techniques, our baselines require data augmentation of image observations to facilitate learning. Specifically, during training, we apply stochastic shifts of 0-3 pixels to both the gripper and static camera images, modifying the original shifts of 0-4 pixels and 0-10 pixels, respectively, as in Yarats \emph{et al.}~\cite{yarats2021drqv2}. Additionally, we apply bilinear interpolation to the shifted images, replacing each pixel with the average of the nearest pixels. This augmentation technique is used consistently across all baselines.

\subsection{7-DoF Action Framework}

In all our experiments, including simulated and real-world scenarios, we employ a 7-dimensional action framework:
\[\left[\delta x, \delta y, \delta z, \delta \phi, \delta \theta, \delta \psi, \textit{gripperAction} \right]\]
The first three dimensions, $\delta x$, $\delta y$, and $\delta z$, represent translations along the $x$, $y$, and $z$ axes, respectively, to control the end-effector's position in 3D space. The following three dimensions, $\delta \phi$, $\delta \theta$, and $\delta \psi$, are Euler angles which specify changes in the end-effector's orientation relative to the robot's base frame. All six of these parameters accept continuous values within the range $[-1, 1]$. The final dimension, \textit{gripperAction}, controls the gripper state. Unlike the others, it accepts discrete values: $-1.0$ for opening and $1.0$ for closing. \ourmodel~treats \textit{gripperAction} as a continuous variable between $-1.0$ and $1.0$. To ensure compatibility with the environment, we apply thresholding before sending this value to the robot. Values greater than 0 trigger an open action, while values less than or equal to 0 cause the gripper to close.

\subsection{Data Collection Details}

\begin{figure}[t]
\centering	\includegraphics[width=1.0\columnwidth]{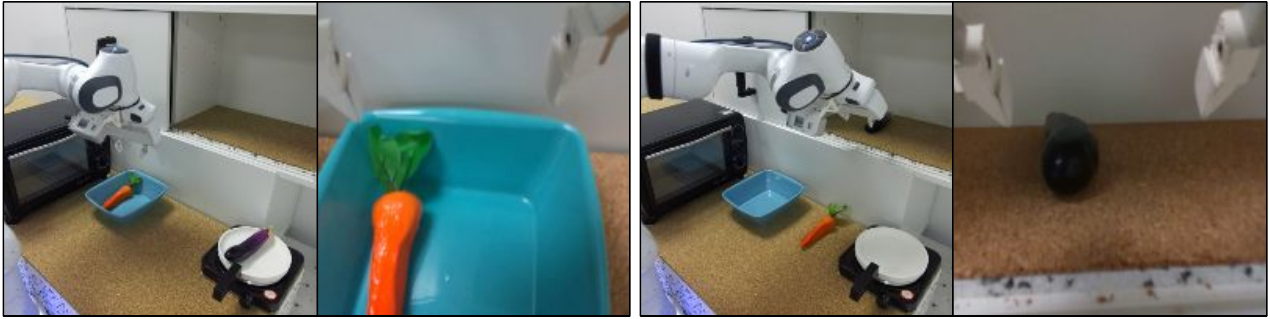}
\caption{Visualization of the real-world environment observations across two time steps. At each time step, the RGB image from the static camera is presented on the left, and the RGB image from the gripper camera is shown on the right.
}
\label{fig:rw_obs}
\end{figure}

For our real-world experiment, we collected three hours of play data by teleoperating a Franka Emika Panda robot with a HTC VIVE Pro headset in a 3D tabletop environment (Figure~\ref{fig:rw_setup}).  The environment contained various everyday objects for interaction, including a stove, a bowl, a cabinet, and toy vegetables like a carrot and an eggplant. During data collection, we recorded a comprehensive set of sensor measurements from the robot. These measurements included proprioceptive data, which provide information about the robot's joint positions and the pose of its end-effector. Additionally, we captured static RGB images ($200 \times 200$ resolution) using an Azure Kinect camera to capture the surrounding environment. We also acquired high-resolution RGB images ($200 \times 200$ resolution) from a FRAMOS Industrial Depth D435e mounted on the robot's gripper, providing detailed information about the objects being manipulated (Figure~\ref{fig:rw_obs}). Lastly, we logged the commanded absolute actions sent to the robot for control purposes.

For training, we derived relative actions by computing the differences in absolute actions between successive time steps. Additionally, we resized the observations to a resolution of \(64 \times 64\). Given the minimal movement between frames, the original recording frequency of 30 Hz was downsampled to 15 Hz.

\end{document}